\documentclass[conference]{IEEEtran}
\IEEEoverridecommandlockouts
\usepackage{cite}
\usepackage{amsmath,amssymb,amsfonts}
\usepackage{algorithmic}
\usepackage{subfigure}
\usepackage{graphicx}
\usepackage{textcomp}
\usepackage{multirow}
\usepackage{xcolor}
\def\BibTeX{{\rm B\kern-.05em{\sc i\kern-.025em b}\kern-.08em
    T\kern-.1667em\lower.7ex\hbox{E}\kern-.125emX}}
\makeatletter 
\newcommand{\linebreakand}{%
  \end{@IEEEauthorhalign}
  \hfill\mbox{}\par
  \mbox{}\hfill\begin{@IEEEauthorhalign}
}
\newcommand\red[1]{{#1}}
\newcommand{\RR}{I\!\!R} 
\usepackage{bbold}
\usepackage{physics}
\usepackage{mathtools}
\usepackage{verbatim}
\usepackage{xr}
\usepackage{stmaryrd}
\usepackage{xcolor}
\usepackage{fancyhdr}

\usepackage{balance}
\usepackage{algorithm}
\usepackage{algorithmic}

\usepackage{placeins}
\usepackage{booktabs}

\definecolor{somegray}{rgb}{0.5, 0.5, 0.5}
\newcommand{\darkgrayed}[1]{\textcolor{somegray}{#1}}
\makeatletter
\newcommand\titleheader[1]{\gdef\@titleheader{#1}}
\AtBeginDocument{%
  \let\st@red@title\@title
  \def\@title{%
    \vskip-8em
    \bgroup\normalfont\large\centering\@titleheader\par\egroup
    \vskip1.0em\st@red@title}
}
\makeatother
\titleheader{\darkgrayed{This paper has been accepted for publication at the\\
IEEE Conference on Artificial Intelligence Circuits and Systems (AICAS), Hangzhou, 2023.
\copyright IEEE}}

\begin{document}




\fancypagestyle{firstpage}{%
  \chead{This paper has been accepted for publication at the\\
IEEE Conference on Artificial Intelligence Circuits and Systems (AICAS), Hangzhou, 2023.
\copyright IEEE}
}
\title{Online Spatio-Temporal Learning with Target Projection}
\author{\IEEEauthorblockN{Thomas Ortner$^{1\dagger\star}$, Lorenzo Pes$^{1,2\dagger}$, Joris Gentinetta$^{1,3}$, Charlotte Frenkel$^{2}$ and Angeliki Pantazi$^{1}$}
\IEEEauthorblockA{\textrm{$^{1}$ IBM Research – Zurich} \\
\textrm{$^{2}$ Microelectronics Department, Delft University of Technology, Delft, The Netherlands}\\
\textrm{$^{3}$ Department of Computer Science, ETH Zurich, Zurich, Switzerland}\\
\textit{$^{\dagger}$ Equal contribution}\\
\textit{$^{\star}$ Correspondence to boh@zurich.ibm.com}
}
}

\maketitle
\thispagestyle{firstpage}

\begin{abstract}
Recurrent neural networks trained with the backpropagation through time (BPTT) algorithm have led to astounding successes in various temporal tasks. However, BPTT introduces severe limitations, such as the requirement to propagate information backwards through time, the weight symmetry requirement, as well as update-locking in space and time. These problems become roadblocks for AI systems where online training capabilities are vital. Recently, researchers have developed biologically-inspired training algorithms, addressing a subset of those problems. In this work, we propose a novel learning algorithm called online spatio-temporal learning with target projection (OSTTP) that resolves all aforementioned issues of BPTT. In particular, OSTTP equips a network with the capability to simultaneously process and learn from new incoming data, alleviating the weight symmetry and update-locking problems. We evaluate OSTTP on two temporal tasks, showcasing competitive performance compared to BPTT. Moreover, we present a proof-of-concept implementation of OSTTP on a memristive neuromorphic hardware system, demonstrating its versatility and applicability to resource-constrained AI devices.
\end{abstract}


\section{Introduction}
Recurrent neural networks (RNNs), a popular type of artificial neural networks (ANNs), have achieved great success in solving temporal tasks such as speech recognition~\cite{Graves2012Nov, Chan2016Mar, Li2021Nov}, translation~\cite{Bahdanau2014Sep} and time series analysis~\cite{Lim2019Dec, Lim2021Apr}, thanks to powerful gradient-based training methods like backpropagation through time (BPTT). However, they still face severe limitations in scenarios where online training is required and cannot be efficiently realized on edge AI devices. In contrast, the human brain solves these tasks online, while consuming low power. When comparing conventional RNNs to biological systems, two striking differences become apparent: the neural dynamics of the units and the employed learning algorithms.

For instance, the neuronal dynamics used in RNNs, mainly based on long short-term memory units (LSTMs) or gated recurrent units (GRUs), are only remotely related to what is observed in the human brain. As an alternative, researchers developed spiking neural networks (SNNs) that aim at reproducing the key dynamics of biological neural cells to take advantage of their sparse and event-driven communication, which is key to the high energy efficiency of the human brain. Recently, SNNs have been incorporated into deep learning frameworks~\cite{neftci2019surrogate, Wozniak2020Jun}, using pseudoderivatives that accommodate for the non-differentiability of the spiking activation function. This enabled SNNs to be trained with gradient-based learning algorithms, such as BPTT, and to achieve performance levels comparable to their conventional RNNs counterparts~\cite{Bohnstingl2022May}.

Furthermore, BPTT raises several concerns regarding its biological plausibility~\cite{Bengio2015Feb, Lillicrap2020Jun}. Firstly, it requires to propagate \textit{information backwards through time} by unrolling the network. Secondly, it uses non-local information to update the synaptic weights. Specifically, it relies on symmetric weights for the forward and the backward pass, leading to the \textit{weight transport problem}, which constrains memory access patterns and leads to inefficient hardware implementations~\cite{Frenkel2021Jun}. Additionally, it requires the update of synaptic weights to occur after all inputs have been processed (\textit{update-locking in time}) and after all the layers of the network computed (\textit{update-locking in space}).

To address these shortcomings, biologically-plausible training algorithms have been developed. Feedback alignment (FA~\cite{Lillicrap2016Nov}), direct feedback alignment (DFA~\cite{Nokland2016Sep})
 solve the weight transport problem. In addition, direct random target projection (DRTP~\cite{Frenkel2021Feb}) projects the target information directly to the individual layers allowing the weight updates to be performed as the forward pass progresses, 
thus solving the update-locking problem in space. 
However, these algorithms have not been designed for RNNs and thus cannot naturally be applied to temporal tasks. To solve the update-locking problem in time, eligibility propagation (e-prop~\cite{Bellec2020}) and online spatio-temporal learning (OSTL~\cite{Bohnstingl_TNNLS2022}) have been introduced. These algorithms allow for online training on temporal tasks while simultaneously processing and learning from new incoming data.

To solve both update-locking problems in space and time, deep continuous local learning (DECOLLE~\cite{Kaiser2020May}) uses local auxiliary loss functions per layer. Furthermore, parallel temporal neural coding network (P-TNCN~\cite{Ororbia2020Jan}) combines the concept of local loss functions with predictive coding. 
However, the use of local loss functions reduces the performance in comparison to BPTT and predictive coding also requires top-down connections. In parallel to our work, the event-based three-factor local plasticity (ETLP~\cite{Quintana2023Jan}) algorithm has been developed, by combining e-prop and DRTP. However, because ETLP is based on e-prop, it cannot be extended to multi-layer RNNs.

In this work, we propose a novel online training algorithm, called online spatio-temporal learning with target projection (OSTTP), which computes the synaptic updates exclusively based on information that is locally available in space and time. OSTTP combines the eligibility traces of OSTL with the target projection signals from DRTP. Our approach is applicable to multi-layer networks and resolves the issues of BPTT by allowing learning while receiving new input data and propagating information only in a forward manner.

For this reason, our approach is a natural match for efficient implementations on edge AI devices. We demonstrate a proof-of-concept realization of OSTTP on memristive in-memory computing neuromorphic hardware (NMHW)~\cite{Khaddam-Aljameh2021Jun}. In particular, our contributions are:
\begin{itemize}
    \item a novel biologically-inspired training algorithm that allows for online training using exclusively locally available information;
    \item an application of OSTTP to spiking neural networks on two challenging temporal tasks showing competitive performance compared to BPTT and outperforming current local learning algorithms;
    \item a proof-of-concept implementation of OSTTP on a neuromorphic hardware-in-the-loop training system.
\end{itemize}

\pagestyle{plain}

\begin{table}[]
    \caption{Comparison of training algorithms from the literature}
    \label{tab:AlgoComp}
    \centering
    \begin{tabular}{l|c|c|c}
         \hline
         \multirow{2}{*}{Algorithm} & No symmetric & No update-locking & No update-locking\\
         & weights & in time & in space\\
         \hline
         BPTT & $\times$ & $\times$ & $\times$\\
         FA & $\surd$ & - & $\times$\\
         DFA & $\surd$ & - & $\times$\\
         DRTP & $\surd$ & - & $\surd$\\
         e-prop & $\surd$ & $\surd$ & $\times$\\
         OSTL & $\times$ & $\surd$ & $\times$\\
         OSTL \textit{rnd} & $\surd$ & $\surd$ & $\times$\\
         DECOLLE & $\surd$ & $\surd$ & $\surd$\\
         ETLP & $\surd$ & $\surd$ & $\surd$\\
         \hline
         \hline
         \textbf{OSTTP} & $\boldsymbol{\surd}$ & $\boldsymbol{\surd}$ & $\boldsymbol{\surd}$\\
         \hline
    \end{tabular}
\end{table}

\section{Neural dynamics and Online Spatio-Temporal Learning with Target Projection}
\begin{figure}[htbp]
  \centering
  \includegraphics[width=1.\columnwidth]{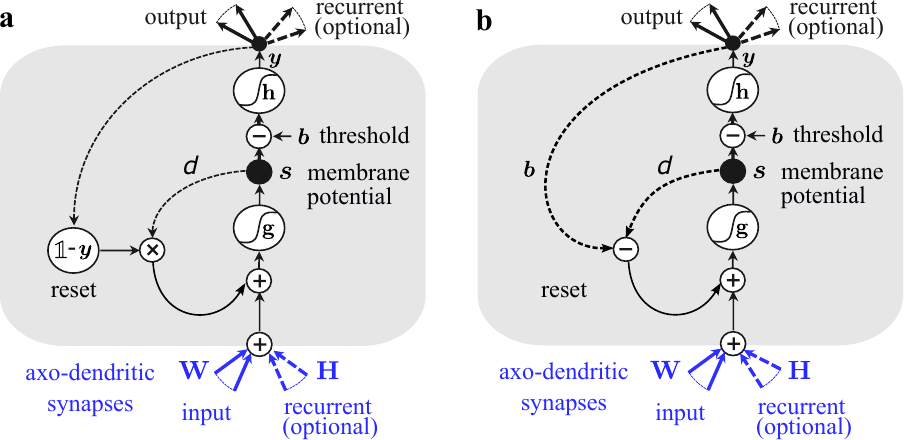}
  \caption{\textbf{Spiking neural units with two different reset mechanisms. Drawings adapted from~\cite{Wozniak2020Jun}.} \textbf{a} Full reset mechanism: in case the neuron emits a spike, the membrane potential is reset to 0. \textbf{b} Soft-reset mechanism: in case the neuron emits a spike, the threshold is subtracted from the membrane potential.}
  \label{fig:Figure1}
\end{figure}

Our biologically-inspired training algorithm is applicable to any neuron model, including RNNs and SNNs. However, we opted to use the spiking neural units (SNUs~\cite{Wozniak2020Jun}) as they leverage the efficient spike-based communication and rich dynamics of biological neurons. The SNU, in its simplest form, implements the ubiquitous leaky integrate-and-fire (LIF) neuron model. Here, we outline the mathematical foundation of the SNU using two slightly different reset mechanisms: a full reset mechanism and a soft-reset mechanism, similar to~\cite{Bellec2020}. Fig.~\ref{fig:Figure1}a shows the SNU with the full reset mechanism, i.e., if the neuron emits a spike, the membrane potential is reset to $0$. The governing equation for $n$ SNUs with the full reset mechanism can be written as:
\begin{align}
    \boldsymbol{s}_{l}^t &= \mathbf{g}\left(\boldsymbol{y}_{0}^t \mathbf{W} + \boldsymbol{y}_{l}^{t-1} \mathbf{H} + d \boldsymbol{s}_{l}^{t-1} \left(\mathbb{1} - \boldsymbol{y}_{l}^{t-1}\right)\right)\\
    \boldsymbol{y}_{l}^t &= \mathbf{h}\left(\boldsymbol{s}_{l}^t-\boldsymbol{b}_{l}\right), \label{SNU}
\end{align}
where $\boldsymbol{s}_l^t \in \RR^{n}$ is the membrane potential of layer $l$ at time $t$, $\boldsymbol{y}_l^t \in \RR^{n}$ is the output of layer $l$ at time $t$, $\boldsymbol{y}_0^t \in \RR^{m}$ signifies the input, $\mathbf{W} \in \RR^{m \times n}$ and $\mathbf{H} \in \RR^{n \times n}$ are the input and the recurrent weight matrices, while $d \in \RR$ is the decay of the membrane potential. $\mathbf{g}$ and $\mathbf{h}$ represent the membrane and output activation functions, with $\mathbf{g}(x)=\mathbb{1}(x)$ and $\mathbf{h}(x)=\boldsymbol{\Theta}(x)$ for all our simulations and experiments. $\boldsymbol{\Theta}$ is the Heaviside step function, i.e. $\boldsymbol{\Theta}\left(\boldsymbol{x}\right) = 1$ if $\boldsymbol{x} > 0$ and $0$ otherwise.

Fig.~\ref{fig:Figure1}b shows the SNU with the soft-reset mechanism, i.e., if the neuron emits a spike, the current threshold is subtracted from the membrane potential, instead of resetting it to zero. The governing equation for $n$ SNUs with the soft-reset mechanism can be written as:
\begin{align}
    \boldsymbol{s}_{l}^t &= \boldsymbol{y}_{0}^t \mathbf{W} + \boldsymbol{y}_{l}^{t-1} \mathbf{H} + d \boldsymbol{s}_{l}^{t-1} - \boldsymbol{y}_{l}^{t-1} \boldsymbol{b}_{l}\\
    \boldsymbol{y}_{l}^t &= \boldsymbol{\Theta} (\boldsymbol{s}_{l}^t-\boldsymbol{b}_{l}). \label{SNUSR}
\end{align}
While the full reset mechanism is more commonly observed in biological systems, the soft-reset mechanism maintains information in the membrane potential, even after a spiking event. Therefore, the soft-reset mechanism is particularly well-suited for temporal tasks that require long time horizons to be solved.

\begin{figure}[htbp]
  \centering
  \includegraphics[width=1.\columnwidth]{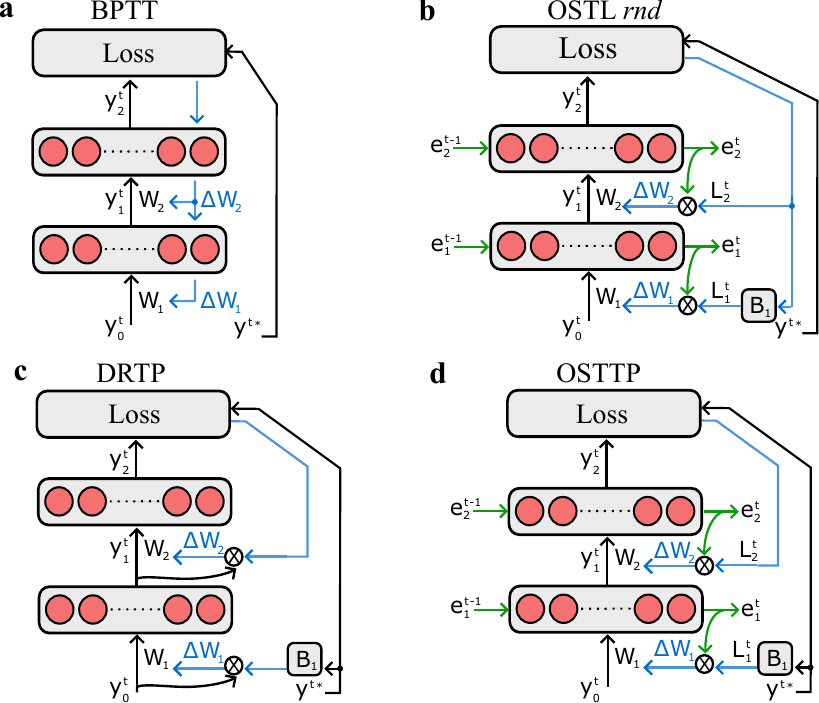}
  \caption{\textbf{Comparison of training algorithms.} \textbf{a} In BPTT, the target vector $\boldsymbol{y}^{t,*}$ is only provided to the loss layer and the parameter updates are computed based on gradient signals traversing through the network architecture in space and time. \textbf{b} In OSTL, the gradient components of BPTT are cleanly separated into temporal gradients, captured by eligibility traces $\mathbf{e}^{t,\boldsymbol{\theta}_l}_{l}$, and into spatial gradients captured by the learning signals $\mathbf{L}^t_{l}$. \textbf{c} In DRTP, the target vector $\boldsymbol{y}^{t,*}$ is directly provided to the individual layers, allowing them to be updated without having to wait for the preceding layer to be computed. \textbf{d} OSTTP combines the eligibility traces from OSTL with the projected target signals from DRTP, making it a fully online training algorithm without any update-locking problems.}
  \label{fig:Figure2}
\end{figure}

In gradient-based algorithms such as BPTT, illustrated in Fig.~\ref{fig:Figure2}a, the update of any model parameter $\boldsymbol{\theta}_l$, for example the input weights $\mathbf{W}$, the recurrent weights $\mathbf{H}$, or the threshold $\boldsymbol{b}$, is performed according to 
\begin{align}
\Delta \boldsymbol{\theta}_l = -\eta \dv{E}{\boldsymbol{\theta}_l} = -\eta \sum_t \dv{E^t}{\boldsymbol{\theta}_l}\label{eq:GenericParameterUpdate},
\end{align}
where $\eta$ is the learning rate and $\dv{E}{\boldsymbol{\theta}_l}$ is the total derivative of the error with respect to the parameter $\theta$ to be updated. As one can see, BPTT requires to send information (i) through layers, starting from the last layer $K$, where the error $E$ is computed, to the layer $l$, as well as (ii) through time, from timestep $t$ until the first timestep $t=0$. For the differentiation of the Heaviside step function, a pseudoderivative is employed, which is described in detail in Section~\ref{sec:results}.

OSTL separates the gradient flow of BPTT into temporal components, traversing only through time within a single layer, and into spatial components, traversing the layers of the network architecture within a single timestep. It introduces eligibility traces $\boldsymbol{e}_l^{t,\theta_l}$ to capture the temporal components and learning signals $\mathbf{L}_l^t$ to capture the spatial components. The parameter update can then be formulated as
\begin{align}
\Delta \boldsymbol{\theta}_l \approx \sum_{t} \mathbf{L_{l}^t}\mathbf{e_{l}^{t,\theta_l}}\label{eq:OSTL},
\end{align}
where the learning signals can be computed in a recurrent form using
\begin{align}
\mathbf{L}^t_{l} &= \mathbf{L}^t_{l+1} \left(\pdv{\boldsymbol{y}^t_{l+1}}{\boldsymbol{s}^t_{l+1}} \pdv{\boldsymbol{s}^t_{l+1}}{\boldsymbol{y}^t_{l}}\right)\label{eq:generalLearningSignalRecursive} \,\, \mathrm{with} \,\, \mathbf{L}^t_{\red{K}} = \pdv{E^t}{\boldsymbol{y}_{\red{K}}^t},
\end{align}
and the eligibility traces as
\begin{align}
\boldsymbol{\epsilon}_l^{t,\boldsymbol{\theta}_l} &= \dv{\boldsymbol{s}_l^t}{\boldsymbol{\theta}_l} = \dv{\boldsymbol{s}_l^{t}}{\boldsymbol{s}_l^{t-1}} \boldsymbol{\epsilon}_l^{t-1,\boldsymbol{\theta}_l} + \left( \pdv{\boldsymbol{s}_l^t}{\boldsymbol{\theta}_l} + \pdv{\boldsymbol{s}_l^t}{\boldsymbol{y}_l^{t-1}} \pdv{\boldsymbol{y}_l^{t-1}}{\boldsymbol{\theta}_l}\right) \label{eq:generalEligibilityVector}\\
\mathbf{e}^{t,\boldsymbol{\theta}_l}_{l} &= \pdv{\boldsymbol{y}^t_{l}}{\boldsymbol{s}^t_{l}} \boldsymbol{\epsilon}^{t,\boldsymbol{\theta}_l}_{l} + \pdv{\boldsymbol{y}^t_{l}}{\boldsymbol{\theta}_{l}}. \label{eq:generalEligibilityTrace}
\end{align}
Note that even though the generic form of the eligibility traces of e-prop appears similar to OSTL, the derivation of the latter maintains gradient equivalence to BPTT for a network with a single recurrent layer and is also applicable to multi-layer RNNs.
More mathematical details and the proofs can be found in~\cite{Bohnstingl_TNNLS2022}.

From the mathematical formulation of the learning signal expressed in~(\ref{eq:generalLearningSignalRecursive}), one can see that $\mathbf{L}^t_{l}$ can only be computed after the preceding layer has been computed, i.e., $\boldsymbol{y}^t_{l+1}$ is needed. Moreover, the learning signal traverses backwards through the layer, e.g., by using $\left(\pdv{\boldsymbol{y}^t_{l+1}}{\boldsymbol{s}^t_{l+1}} \pdv{\boldsymbol{s}^t_{l+1}}{\boldsymbol{y}^t_{l}}\right)$, and thus requiring symmetric weights $W_{l+1}$ for the forward as well as for the backward passes. Therefore, it is evident that both the update-locking problem in space as well as the weight transport problem are present. To solve the latter, OSTL $\textit{rnd}$ has been introduced~\cite{Bohnstingl_TNNLS2022}, pairing OSTL with DFA. In OSTL $\textit{rnd}$, illustrated in Fig.~\ref{fig:Figure2}b, the learning signals for each layer are computed as follows, using a fixed random feedback matrix $\mathbf{B}_l$
\begin{align}
\mathbf{L}^t_{l} &= \begin{cases}
\dv{E^t}{\boldsymbol{\theta}_l} &l = K\\
\mathbf{B}_l \dv{E^t}{\boldsymbol{\theta}_l} &l \le K.
\end{cases}
\end{align}
While OSTL \textit{rnd} addresses the weight transport problem, it still leaves the update-locking problem in space unsolved. DRTP resolves this issue by directly propagating the target vector $\boldsymbol{y}^{t,*}$ to the loss function, as well as to the individual layers, also using a fixed random matrix $\mathbf{B}_l$, see Fig.~\ref{fig:Figure2}c. In this work we combine OSTL and DRTP, resulting in our proposed training algorithm OSTTP, see Fig.~\ref{fig:Figure2}d.
While the eligibility traces of OSTL remain unchanged, the learning signal for this algorithm can be computed as
\begin{align}
\mathbf{L}^t_{l} &= \begin{cases}
\dv{E^t}{\boldsymbol{\theta}_l} &l = K\\
\mathbf{B}_l \boldsymbol{y}^{t,*} &l \le K.
\end{cases}
\end{align}
As one can see, the learning signals for all layers of the network can be computed separately, without waiting for the preceding layer to be computed. Therefore, OSTTP resolves the update-locking problems, both in time and in space. 
Tab.~\ref{tab:AlgoComp} summarizes the major characteristics of the individual training algorithms from the literature.

\section{Results}\label{sec:results}
\subsection{Software simulations}
The OSTTP performance was evaluated on two temporal datasets: The Johann Sebastian Bach (JSB) music chorales~\cite{Boulanger-Lewandowski2012Jun} and the Spiking Heidelberg Digits (SHD)~\cite{Cramer2020Dec}. 

\begin{figure}[htbp]
  \centering
  \includegraphics[width=1.\columnwidth]{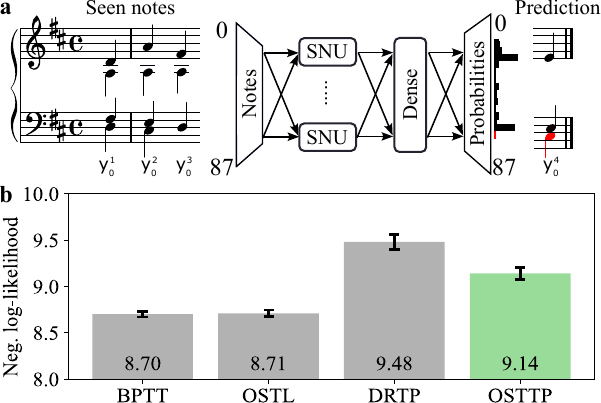}
  \caption{\textbf{Music prediction task using the JSB dataset.} \textbf{a} Network architecture comprising an SNU layer with the full reset mechanism and a dense output layer with a sigmoidal activation function. The inputs to the network are time sequences of binary encoded vectors, representing the notes played on the 88 keys of a piano. The network takes these binary-encoded inputs for timesteps $t=0$ to $t$ and produces a probability distribution of the 88 keys that will be played at timestep $t+1$. We used the standard split for this dataset consisting of 229 music pieces for training and 77 pieces for testing. \textbf{b} Performance comparison of the different training algorithms.}
  \label{fig:Figure3}
\end{figure}

For the first dataset, the objective at every timestep $t$ is to predict the musical notes of a piano that will be played at timestep $t+1$, based on the already played notes until timestep $t$. To solve this task, we employed a single feedforward layer of $150$ SNUs, using the full reset mechanism, and a dense readout layer with $88$ units and a sigmoidal activation function. Fig.~\ref{fig:Figure3}a illustrates the network architecture as well as its inputs and outputs. For evaluation, we use the negative log-likelihood metric, where a lower value indicates a better prediction of the network. The SNUs were configured with trainable parameters $\boldsymbol{\theta}=\{\mathbf{W}, \mathbf{H}, \boldsymbol{b}\}$, a decay of $d=0.4$ for BPTT, $d=0.5$ for OSTL, $d=0.5$ for DRTP and $d=0.6$ for OSTTP, as well as with the pseudoderivative $\dv{\mathbf{\Theta}(\boldsymbol{x})}{\boldsymbol{x}}=\mathbb{1}-\mathbf{tanh}^2(\boldsymbol{x})$. Furthermore, we employed a standard stochastic gradient descent optimizer (SGD) with learning rates of $0.001$ for BPTT, $0.0005$ for OSTL, $0.0005$ for DRTP and $0.0005$ for OSTTP. As one can see in Fig.~\ref{fig:Figure3}b, OSTL maintains gradient equivalence to BPTT for this network architecture. However, both algorithms suffer from update-locking problems. When using DRTP, modified for a temporal task where the parameter updates are accumulated over time, the update-locking problem in space is resolved, but the performance of the network deteriorates. This is because, in contrast to the gradient signals in BPTT or in OSTL, the target signals used to compute the parameter updates at timestep $t$ do not contain information from previous timesteps. On the contrary, OSTTP combines the benefits from the traces of OSTL, with the learning signals from DRTP, thereby resolving the update-locking problem of OSTL and at the same time lowering the performance gap between DRTP and BPTT.

\begin{figure}[htbp]
  \centering
  \includegraphics[width=1.\columnwidth]{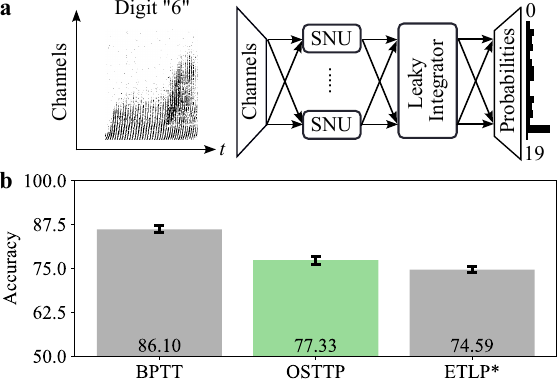}
  \caption{\textbf{Temporal classification task using the SHD dataset.} \textbf{a} Network architecture comprising an SNU layer with the soft-reset mechanism and a leaky integrating output layer. The inputs to the network are spike trains, representing recordings from speakers uttering digits from zero to nine in both German and English. The original audio recordings were converted into spike trains, using an artificial cochlear model. The dataset includes $12$ speakers and a total number of $10420$ samples and we used the standard split of $8156$ examples for training and $2264$ for testing. \textbf{b} Performance comparison of the different training algorithms. *ETLP~\cite{Quintana2023Jan} developed in parallel to this work.}
  \label{fig:Figure4}
\end{figure}

For the SHD dataset, the objective is to classify spoken digits. We employed a single recurrent layer of $450$ SNUs, using the soft-reset mechanism, and a leaky-integrating output layer of $20$ units. Fig.~\ref{fig:Figure4}a illustrates the network architecture as well as its inputs and outputs. The SNUs were configured with trainable parameters $\boldsymbol{\theta}=\{\mathbf{W}, \mathbf{H}\}$, a decay of $d=0.83$ for BPTT, $d=0.95$ for OSTTP, as well as with the pseudoderivative $\dv{\mathbf{\Theta}(\boldsymbol{x})}{\boldsymbol{x}}= \frac{1}{(100 \lvert \boldsymbol{x} \rvert + 1)^2}$. The leaky integrator was configured with a trainable parameter $\theta=\{\mathbf{W}\}$, with a decay of $\tau=0.98$ for BPTT and of $\tau=0.99$ for OSTTP. Furthermore, we employed the ADAM optimizer with learning rates of $0.00035$ for BPTT and  $0.0002$ for OSTTP. For evaluation we employed the common accuracy metric, which we computed as the maximum testing accuracy over $200$ training epochs, averaged over 5 random seeds. The results are shown in Fig.~\ref{fig:Figure4}b, where OSTTP achieves an accuracy of $(77\pm0.8)\%$ accuracy. Furthermore, its performance surpasses the one of ETLP, despite ETLP using threshold adaptive neurons that increase the computational complexity. Note that network performance degrades to $(75\pm0.8)\%$ when using the full reset mechanism, because the long-term temporal dependencies cannot be captured.

\subsection{Demonstration on neuromorphic hardware}
\begin{figure}[htbp]
  \centering
  \includegraphics[width=1.\columnwidth]{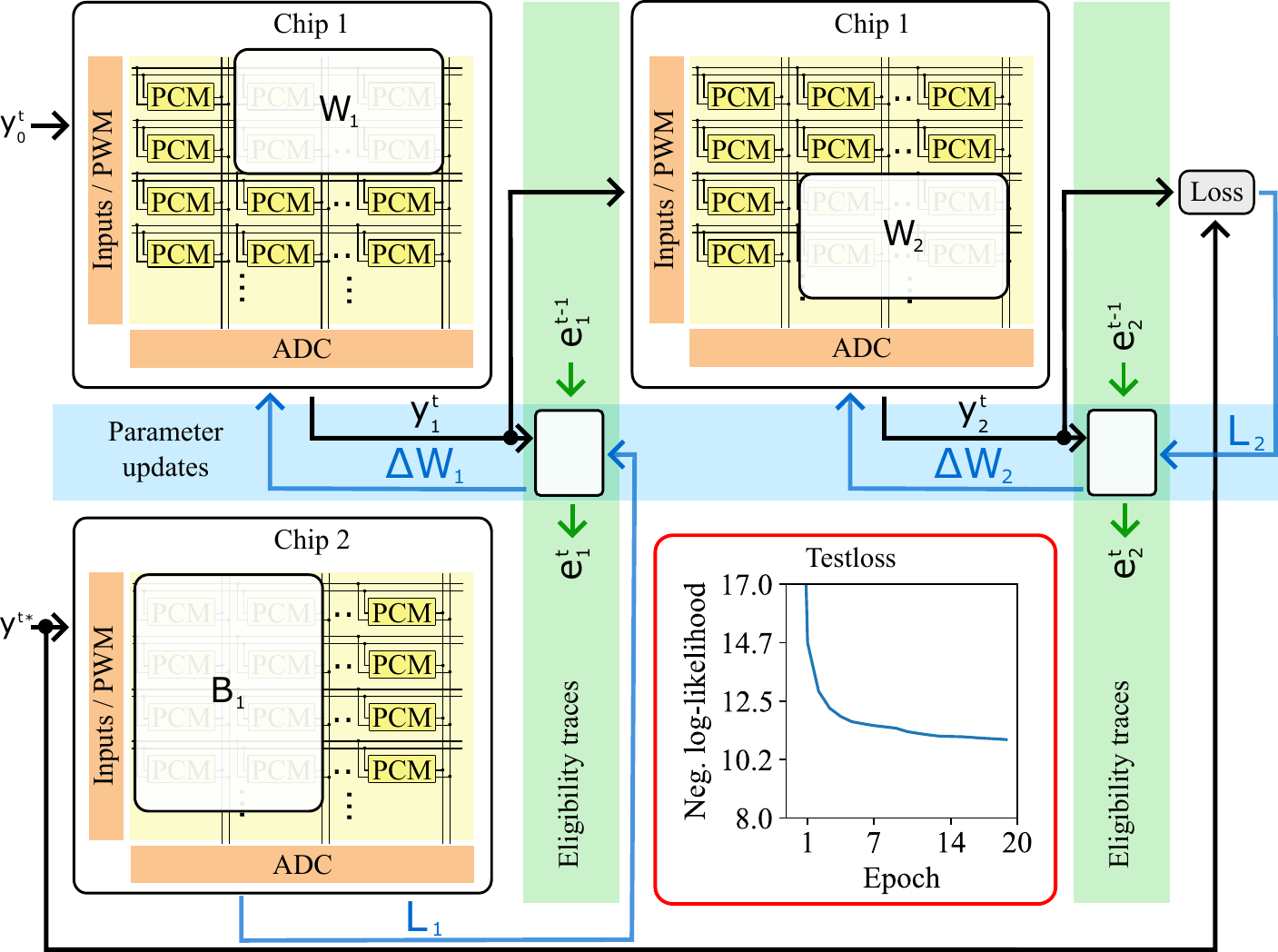}
  \caption{\textbf{Proof-of-concept demonstration of OSTTP on NMHW.} Two chips with crossbar arrays implement the network architecture used for the JSB task. The weights of the SNU layer $\mathbf{W}_1$ and of the dense layer $\mathbf{W}_2$ are placed on chip 1, while the fixed random matrix $\mathbf{B}$ is placed on chip 2. The eligibility traces are kept in software and the computed parameter updates are applied to the NMHW in the form of individual electrical pulses. The red inset shows the test loss evolution over epochs.}
  \label{fig:Figure5}
\end{figure}

We also demonstrate the viability of OSTTP for a neuromorphic hardware system. To this end, we employ an in-memory computing platform~\cite{Khaddam-Aljameh2021Jun}, based on memristive phase-change memory (PCM) devices~\cite{Sebastian2020Jul}. These devices belong to a special class of emerging materials that can represent information using their conductance values $G$ and simultaneously perform computations within the memory elements. The particular neuromorphic hardware system used in this work is comprised of two chips, each hosting a crossbar arrangement of 256x256 unit cells, consisting of four PCM devices in a differential configuration~\cite{Khaddam-Aljameh2021Jun}.
We leveraged this setup to map the synaptic weights $\mathbf{W}_1$ of the SNU layer, configured with $d=0.6$ and $\dv{\mathbf{\Theta}(\boldsymbol{x})}{\boldsymbol{x}}=\mathbb{1}-\mathbf{tanh}^2(\boldsymbol{x})$, 
as well as $\mathbf{W}_2$ of the dense output layer onto one chip. To parallelize the learning signal computation, the fixed random matrix $\mathbf{B}$ was mapped on the second chip. Each element of these matrices is represented by four PCM devices as $W_{\mathrm{ij}} = ((G_{\mathrm{A}}^+ + G_{\mathrm{B}}^+) - (G_{\mathrm{A}}^- + G_{\mathrm{B}}^-)) / 2$. Fig.~\ref{fig:Figure5} shows an illustration of the realization of OSTTP on the NMHW. The weights are mapped onto the NMHW, which is used to perform the matrix multiplications, while the eligibility traces are calculated in software. In order to change the synaptic weights, electrical pulses are sent to the PCM devices, either increasing or decreasing their conductance levels. We used this hardware-in-the-loop setup, with SGD using a learning rate of $0.001$ which we multiplied by $0.6$ every 5 epochs, to train our network and achieved a negative log-likelihood of $11.0\pm0.2$ for the JSB task. 
Note that the network architecture used to solve the SHD dataset was beyond the size capabilities of the experimental NMWH.

\section{Discussion}
In this paper, we introduced a novel biologically-inspired training algorithm called online spatio-temporal learning with target projection. This algorithm solves several shortcomings of BPTT, in particular the weight transport problem, the requirement to send information backwards through time and the update-locking problems, both in time and space. It enables a network to be trained fully online, exclusively with information that is locally available to the individual parameters. Although OSTTP has been formulated to accumulate weight updates in time before applying them to the parameters, which would make it non-local in time (see (\ref{eq:OSTL})), there are approaches to alleviate this by applying the parameter updates directly at every timestep~\cite{Bohnstingl_TNNLS2022, Frenkel2022Feb}. 
Additionally, we demonstrated the performance of OSTTP on two different temporal tasks and achieved performance levels close to the BPTT baseline. Moreover, we showcased the applicability of our algorithm on a PCM-based neuromorphic hardware system, using a hardware-in-the-loop setting. This demonstration highlights the main advantages of OSTTP: firstly, it does not require symmetric weights, which cannot be used with the specific NMHW, and secondly, the learning signals of OSTTP for all layers can be computed in parallel to the forward pass because only local information is used. This makes OSTTP a perfect match for low-latency massively-parallel NMHW systems operating in a fully forward manner.
 
\section*{Acknowledgment}
We thank the In-Memory Computing team at IBM for their technical support with the PCM-based NMHW as well as the IBM Research AI Hardware Center. This project has received funding from the ERA-NET CHIST-ERA-18-ACAI-004 programme by SNSF under the project number 20CH21\_186999 / 1.

\balance
\bibliographystyle{./bibliographies/IEEEtran}
\bibliography{./Bibliography.bib}
\end{document}